%% file: main.tex
\documentclass{article}

  \usepackage[dblblindworkshop, final]{neurips_2025}
\workshoptitle{The 3rd Workshop on Imageomics: Discovering Biological Knowledge from Images Using AI}

\input{preamble}

\usepackage[utf8]{inputenc} %
\usepackage[T1]{fontenc}    %
\usepackage{hyperref}       %
\usepackage{url}            %
\usepackage{booktabs}       %
\usepackage{amsfonts}       %
\usepackage{nicefrac}       %
\usepackage{microtype}      %
\usepackage{xcolor}         %
\usepackage{graphicx}
\usepackage{multirow}
\usepackage{enumitem}

\usepackage{adjustbox}
\usepackage{titlesec}  %

\titlespacing*{\paragraph}   %
  {0pt}                      %
  {2ex plus 0.5ex minus 0.2ex} %
  {0.25em}                      %
\title{BarcodeMamba+: Advancing State-Space Models\\
for Fungal Biodiversity Research}

\author{%
  \normalfont
  \textbf{Tiancheng Gao}\textsuperscript{1,2}\quad
  \textbf{Scott C.~Lowe}\textsuperscript{2}\quad
  \textbf{Brendan Furneaux}\textsuperscript{3}\quad
  \textbf{Angel X.~Chang}\textsuperscript{4,5}\\
  \textbf{Graham W.~Taylor}\textsuperscript{1,2}\thanks{Author for correspondence: \texttt{gwtaylor@uoguelph.ca}} \\
  \\
  \textsuperscript{1}University of Guelph \quad
  \textsuperscript{2}Vector Institute \quad
  \textsuperscript{3}University of Jyväskylä \\
  \textsuperscript{4}Simon Fraser University \quad
  \textsuperscript{5}Amii
}

\begin{document}

\maketitle

\begin{abstract}
    Accurate taxonomic classification from DNA barcodes is a cornerstone of global biodiversity monitoring, yet fungi present extreme challenges due to sparse labelling and long-tailed taxa distributions. Conventional supervised learning methods often falter in this domain, struggling to generalize to unseen species and to capture the hierarchical nature of the data. To address these limitations, we introduce BarcodeMamba+, a foundation model for fungal barcode classification built on a powerful and efficient state-space model architecture. We employ a pretrain and fine-tune paradigm, which utilizes partially labelled data and we demonstrate this is substantially more effective than traditional fully-supervised methods in this data-sparse environment. During fine-tuning, we systematically integrate and evaluate a suite of enhancements---including hierarchical label smoothing, a weighted loss function, and a multi-head output layer from MycoAI---to specifically tackle the challenges of fungal taxonomy. Our experiments show that each of these components yields significant performance gains. On a challenging fungal classification benchmark with distinct taxonomic distribution shifts from the broad training set, our final model outperforms a range of existing methods across all taxonomic levels. Our work provides a powerful new tool for genomics-based biodiversity research and establishes an effective and scalable training paradigm for this challenging domain. Our code is publicly available at \url{https://github.com/bioscan-ml/BarcodeMamba}.
\end{abstract}

\section{Introduction}
DNA barcodes, short standardized DNA sequences used for specimen recognition and species identification, enable large-scale, automated biodiversity monitoring~\citep{barcode}.
Fungal biodiversity presents an extreme challenge for barcode classification. Visual and morphological features help identify other taxa, but fungal species identification is often confounded by minimalistic features, necessitating an almost complete reliance on DNA sequences~\citep{bickford2007cryptic}.
Currently, up to 93\% of collected fungal samples remain unannotated at the species level~\citep{mycoai}.

This annotation sparsity has exposed fundamental limitations in existing computational approaches. Traditional algorithmic methods like BLAST~\citep{blast}, RDP classifier~\citep{wang2007naive}, and dnabarcoder~\citep{vu2022dnabarcoder} are standard tools for sequence identification but face prohibitive inference times on large datasets and poor generalization to novel taxa.
Learning-based methods using specialized convolutional neural networks (CNNs) and transformer architectures show promise with fully supervised training \citep{cnnencoder,mycoai} but require densely labelled data, making them vulnerable to the class imbalance and label sparsity that characterize fungal datasets.

Foundation models tackle sparse training labels through a pretrain + fine-tune paradigm. The vast amounts of unlabelled data can be harnessed during pretraining to learn rich, generalizable representations, before adapting to specific tasks with fine-tuning.
This approach is effective for biodiversity applications where unlabelled data vastly outnumbers annotated specimens, as demonstrated by the transformer-based BarcodeBERT~\citep{BarcodeBERT} and BarcodeMAE~\citep{barcodemae}.

State-space models (SSMs)~\citep{s4} and particularly the Mamba architecture \citep{mamba,mamba2}, have emerged as compelling alternatives to Transformers for sequence modelling. SSMs offer competitive performance with significantly lower computational overhead, making them attractive for large-scale biodiversity applications where datasets contain millions of sequences.
Our previous work \citep{barcodemamba} introduced BarcodeMamba, demonstrating the effectiveness of SSMs for insect barcode (COI) classification.
This suggests a strong potential application of SSMs to fungal data, which faces similar challenges.

We introduce \textbf{BarcodeMamba+}, which adapts BarcodeMamba for hierarchical fungal ITS barcode classification.
Our experiments demonstrate BarcodeMamba+ outperforms established methods across taxonomic ranks on standard fungal classification benchmarks.
Our contributions are:%
\begin{enumerate}%
    \item The development and comprehensive evaluation of BarcodeMamba+, an SSM-based foundation model for fungal barcode classification.
    \item Demonstration that pretrain + fine-tune approaches outperform fully-supervised methods in this annotation-sparse, taxonomically diverse domain.
    \item Systematic analysis of hierarchical smoothing, inverse square root weighted loss (hereafter shortened to weighted loss), and multi-head outputs for adapting foundation models to hierarchical taxonomic classification.
    \item Analysis of model scaling effects on taxonomic classification performance.
\end{enumerate}
\section{Methods}
\subsection{Dataset}

Our experiments use the MycoAI \citep{mycoai} splits of the UNITE+INSD data~\citep{unite}, a comprehensive fungal internal transcribed spacer (ITS) barcode repository. 
\paragraph{Training and Validation Sets.}
The training set is comprised of 5.23\,M sequences, representing 14.7\,k distinct species across a taxonomic hierarchy of 18 phyla, 70 classes, 231 orders, 791 families, and 3,695 genera. Only 7\% of the samples are annotated to species-level. This creates a complex multi-label, hierarchical classification challenge. The validation set contains 10.5\,k sequences, randomly sampled from~\citet{unite}.

\paragraph{Test Sets.}
We use the three MycoAI test sets, representing distinct taxonomic distribution shifts from the broad training set. \autoref{s:dataset-analysis} analyzes the species-level and identical-barcode overlap between the training set and each test set. Test examples belonging to classes that were unobserved during training were omitted from our evaluation.
\begin{itemize}
    \item \textbf{Test Set 1: Yeast}~\citep{test1}. Contains 4.4\,k ITS sequences from yeast species, evaluating the model's generalization to a specific and taxonomically concentrated clade.
    \item \textbf{Test Set 2: Filamentous Fungi}~\citep{test2}. A set of 11.6\,k sequences from filamentous fungi, a broad but distinct collection of taxa not necessarily well-represented in the training set.
    \item \textbf{Test Set 3: MycoAI Benchmark}~\citep{mycoai}. The largest test set with 367\,k samples, serving as a comprehensive benchmark for overall performance and robustness.
\end{itemize}

\paragraph{Data Preprocessing.}
We used the preprocessed MycoAI dataset \citep{mycoai_data}, with four filtering steps: (1) removal of duplicate sequence-label pairs, (2) exclusion of sequences with length more than four standard deviations from the mean (558.0\,bp $\pm$ 126.2\,bp), (3) removal of sequences with over 5\% ambiguous bases, and (4) elimination of taxonomic classes with fewer than three representative samples. The remaining sequences in the training split are annotated to varying depths within the seven-level taxonomic hierarchy (kingdom, phylum, class, order, family, genus, and species). The dataset was partitioned into training, validation, and test splits after all the above preprocessing steps.
\subsection{Model Architectures}
\subsubsection{Baselines}
We compare BarcodeMamba+ against baselines from three categories. BLAST~\citep{blast} serves as a representative non-learning algorithmic method. For fully-supervised deep learning models, we compare against a CNN Encoder \citep{cnnencoder}, and both MycoAI-CNN and MycoAI-BERT \citep{mycoai}. BarcodeBERT~\citep{BarcodeBERT} provides a competitive foundation model baseline, pretrained on COI barcodes. The two MycoAI models incorporate the same enhancements for hierarchical modelling which we evaluate in \autoref{s:ablation-b}. Complete architectural details and experimental configurations for all baselines are provided in \appref{model_architecture_details_baseline}.
\subsubsection{BarcodeMamba+}
Our BarcodeMamba+ model adapts the BarcodeMamba SSM architecture for hierarchical fungal ITS barcode classification. We use a BPE tokenizer following~\citet{mycoai}'s recommendation for fungal data. Complete implementation details are provided in \appref{model_architecture_details_barcodemambaplus}. 
\paragraph{Training Paradigm.}
We employ a two-stage approach:
\begin{compactitem}
    \item \textbf{Pretraining:} The tokenizer and model learn fungal ITS sequence patterns from unlabelled UNITE+INSD data through next-token prediction, without taxonomic labels.
    \item \textbf{Fine-tuning:} We add a classification head and fine-tune on labelled data, incorporating the enhancements from \appref{sec:enhancements} to address hierarchical labels and class imbalance.
\end{compactitem}
\section{Experiments}
\begin{table}[tb]
    \centering
    \caption{Performance of BarcodeMamba+ and baselines on the three test sets for taxonomic ranks family, genus, and species. We report accuracy (micro) (\%), model size (parameters), and inference time per sample (ms). $\uparrow$: higher is better; $\downarrow$: lower is better. Bold: \best{best}; underlined: \runnerup{second best}.}
    \begin{adjustbox}{width=\textwidth}
    \small
    \begin{tabular}{@{}lccccccccccr@{}}
    \toprule
     & \multicolumn{3}{c}{Yeast Acc. (\%)$\uparrow$} & \multicolumn{3}{c}{Filamentous Acc. (\%)$\uparrow$} & \multicolumn{3}{c}{MycoAI Acc. (\%)$\uparrow$} &  &  \\ \cmidrule(l){2-4}\cmidrule(l){5-7}\cmidrule(l){8-10}
    Model & Fam. & Gen. & Sp. & Fam. & Gen. & Sp. & Fam. & Gen. & Sp. & Size $\downarrow$ & Time $\downarrow$ \\ \midrule
    BLAST & 86.6 & 92.9 & 75.4 & 81.4 & 71.5 & 33.4 & 94.7 & 93.1 & 55.0 & \na & 208.6\,ms \\ \midrule
    MycoAI-CNN (Vu) & 90.5 & 86.4 & 60.0 & 84.1 & 69.8 & 28.2 & 93.9 & 87.8 & 57.1 & 11.6\,M & 11.8\,ms \\
    MycoAI-BERT (base) & 88.9 & 75.7 & 33.5 & 85.1 & 60.8 & 16.6 & 93.2 & 80.3 & 39.3 & 18.4\,M & 4.5\,ms \\
    CNN Encoder & 94.1 & 88.3 & 67.6 & 84.5 & 69.1 & 31.4 & 97.5 & 93.6 & 72.6 & 12.1\,M & 5.8\,ms \\ \midrule
    BarcodeBERT & 95.4 & 88.6 & 59.1 & 87.8 & 70.2 & 27.7 & 97.8 & 92.0 & 58.9 & 44.6\,M & 8.8\,ms \\
    BarcodeMamba+ & \runnerup{98.7} & \runnerup{95.3} & \runnerup{80.6} & \best{92.6} & \runnerup{81.1} & \runnerup{46.5} & \runnerup{99.0} & \runnerup{96.5} & \runnerup{81.7} & 12.1\,M & 8.0\,ms \\
    BarcodeMamba+ (large) & \best{98.8} & \best{95.9} & \best{83.6} & \runnerup{92.5} & \best{81.6} & \best{50.4} & \best{99.3} & \best{97.7} & \best{88.9} & 49.2\,M & 14.7\,ms \\ \bottomrule
    \end{tabular}
    \end{adjustbox}
\label{tab:results-main}
\end{table}
\subsection{Comparison study}
We trained the models (as described in \appref{sec:implementation-details}), then evaluated the performance on the three test datasets. The results (\autoref{tab:results-main}) demonstrate BarcodeMamba+ outperforms all baseline models across all taxonomic levels and metrics, while maintaining inference efficiency. On the largest MycoAI test set (Benchmark), our model achieves a species-level accuracy of 81.7\%, surpassing the next-best performing baseline, CNN Encoder, by 9.1 percentage points (72.6\%). This gap is even more pronounced on the challenging Filamentous Fungi test set, where our model's species-level accuracy (46.5\%) is over 15 points higher than that of CNN Encoder (31.4\%). This highlights our architecture's enhanced robustness to the distributional shifts present between the training domain and the Filamentous Fungi test set. 

Our model achieves this performance with a compact model size of 12.1\,M parameters, comparable to MycoAI-CNN (Vu) (11.6\,M) and MycoAI-BERT (base) (18.4\,M) and significantly smaller than BarcodeBERT (44.6\,M). Compared to the non--learning-based baseline BLAST, our model achieves vastly higher accuracy on fine-grained classification (e.g., 81.7\% vs.~55.0\% on MycoAI species) and demonstrates over 25\texttimes\ faster inference (8.0\,ms vs.~208.6\,ms), rendering it far more suitable for large-scale biodiversity applications. After scaling up to 49.2\,M parameters, BarcodeMamba+ improves performance on every task. It boosts the species-level accuracy on MycoAI by another 7.2 points, from an already high 81.7\% to an exceptional 88.9\%. Similarly, on Filamentous Fungi, Species accuracy increases from 46.5\% to 50.4\%. This result confirms that our model architecture scales effectively, and its capacity to leverage increased parametrization translates directly into improved accuracy, especially for classifying the long tail of rare species. 

\subsection{Ablation Studies}
We conduct two ablation studies, with the first study focusing on the effectiveness of pretraining and different tokenizers, and the second on the impact of the three enhancements for hierarchical data.
\subsubsection{Ablation A: Pretrain + Finetune vs.~Supervised Learning on UNITE Dataset}
We compare pretrain + fine-tune against fully supervised training while evaluating three tokenization methods. All models use hierarchical label smoothing, multi-head outputs, and weighted loss. Results, shown in \autoref{tab:ablation-pretrain-tokenizer}, demonstrate the benefits of pretraining and find BPE is the best tokenizer.\looseness=-1
\begin{table}[t]
    \centering
    \caption{Ablation comparing three tokenizers and two training paradigms: supervised from scratch (\incorrectmark) and fine-tuned following pretraining (\correctmark). Results show accuracy (micro), precision (macro), and recall (macro) on three test sets at family, genus, and species level. Bold: \best{best} result for a given taxonomic rank and test set; underlined: \runnerup{second best}.}
    \small
    \begin{tabular}{@{}llcccccccccc@{}}
    \toprule
     &  &  & \multicolumn{3}{c}{Accuracy (\%)$\uparrow$} & \multicolumn{3}{c}{Precision (\%)$\uparrow$} & \multicolumn{3}{c}{Recall (\%)$\uparrow$} \\ \cmidrule(l){4-6} \cmidrule(l){7-9} \cmidrule(l){10-12}
     Test set & Tokenizer & Pretrain & Fam. & Gen. & Sp. & Fam. & Gen. & Sp. & Fam. & Gen. & Sp. \\ \midrule
     Yeast & Char & \correctmark & 98.4 & 94.6 & 77.4 & 90.5 & 88.7 & \runnerup{72.9} & 87.0 & 79.5 & 39.7 \\
     &  & \incorrectmark & 97.9 & 94.2 & 76.7 & 86.9 & 87.1 & 72.0 & 87.8 & 77.9 & 39.6 \\
     & k-mer & \correctmark & \runnerup{98.6} & \runnerup{94.8} & 77.8 & \best{93.2} & \runnerup{90.5} & 72.6 & \best{90.5} & \runnerup{79.8} & 40.9 \\
     &  & \incorrectmark & 97.8 & 93.9 & 73.0 & 84.3 & 83.1 & 64.9 & 83.7 & 73.4 & 32.3 \\
     & BPE & \correctmark & \best{98.7} & \best{95.3} & \best{80.6} & \runnerup{93.1} & \best{92.3} & \best{77.0} & \runnerup{90.2} & \best{82.2} & \best{46.0} \\
     &  & \incorrectmark & 97.9 & 93.7 & \runnerup{78.6} & 79.4 & 84.6 & 72.0 & 87.0 & 79.0 & \runnerup{42.6} \\ \midrule
     Filamentous & Char & \correctmark & \runnerup{91.7} & 79.5 & 42.2 & 80.5 & \runnerup{68.2} & \runnerup{44.2} & \runnerup{75.3} & 56.2 & 26.1 \\
     &  & \incorrectmark & 91.4 & \runnerup{79.7} & 42.0 & 79.8 & 67.2 & 43.5 & 74.5 & 55.4 & 25.9 \\
     & k-mer & \correctmark & 91.4 & 78.9 & 42.3 & \runnerup{80.9} & 65.8 & 42.2 & 74.6 & 53.8 & 25.9 \\
     &  & \incorrectmark & 89.3 & 74.6 & 36.4 & 73.4 & 58.2 & 34.3 & 67.3 & 45.9 & 20.8 \\
     & BPE & \correctmark & \best{92.6} & \best{81.1} & \best{46.5} & \best{81.8} & \best{71.2} & \best{48.9} & \best{77.3} & \best{60.7} & \best{31.3} \\
     &  & \incorrectmark & 90.3 & 78.6 & \runnerup{43.2} & 75.9 & 66.7 & 43.6 & 73.5 & \runnerup{57.0} & \runnerup{27.3} \\ \midrule
     Myco & Char & \correctmark & 98.8 & 96.2 & 79.0 & 95.6 & 91.1 & 84.8 & 95.8 & 89.9 & 54.5 \\
     &  & \incorrectmark & 98.7 & 95.9 & 78.2 & 94.2 & 90.6 & 85.2 & 95.6 & 89.5 & 55.0 \\
     & k-mer & \correctmark & \runnerup{99.0} & \best{96.9} & \runnerup{81.1} & \best{96.4} & \best{93.2} & \runnerup{86.4} & \best{97.7} & \best{93.2} & 56.5 \\
     &  & \incorrectmark & \runnerup{99.0} & \runnerup{96.5} & 77.0 & \runnerup{96.1} & \runnerup{92.6} & 82.2 & 96.6 & 89.1 & 45.6 \\
     & BPE & \correctmark & \best{99.0} & \runnerup{96.5} & \best{81.7} & 95.2 & 91.3 & \best{88.3} & \runnerup{97.0} & \runnerup{93.0} & \best{65.8} \\
     &  & \incorrectmark & 98.8 & 95.8 & 78.8 & 93.5 & 89.0 & 85.7 & 96.1 & 89.9 & \runnerup{57.5} \\ \bottomrule
    \end{tabular}
    \label{tab:ablation-pretrain-tokenizer}
\end{table}

\subsubsection{Ablation B: Label smoothing, Multi-head, and Weighted loss}
\label{s:ablation-b}
Various methods for handling hierarchical labels present opportunities to improve model training, which we incorporated into our model training paradigm.
To investigate the impact of each of these on the performance of our model, we ablated these configurations, with results shown in \autoref{tab:ablation-smoothing} and statistical tests shown in \appref{s:statistical-tests}.
We find that hierarchical label smoothing consistently improves performance across all metrics (avg. $+3.3$\% acc.), whereas standard smoothing does not provide a significant benefit. Using a weighted loss also provides a consistent improvement for the imbalanced data (avg. $+4.1$\% acc.). However, multi-head outputs provided inconsistent gains over using a single, species-level, head (avg. $-0.04$\% acc.) when fine-tuning on data labelled to species.

\begin{table}[t]
    \centering
    \caption{Ablation of supervised learning enhancements: label smoothing (None, Standard, Hierarchical), weighted loss (WL), and multi-head outputs (MH). \correctmark{}: enabled, \incorrectmark{}: ablated (standard alternative). Results show accuracy, precision, and recall at the species level across test sets. Bold: \best{best} result; underlined: \runnerup{second best}.\looseness=-1}
    \begin{adjustbox}{width=\textwidth}
    \small
    \begin{tabular}{@{}lccccccccccc@{}}
        \toprule
         \multicolumn{3}{c}{Components} & \multicolumn{3}{c}{Accuracy (\%)$\uparrow$} & \multicolumn{3}{c}{Precision (\%)$\uparrow$} & \multicolumn{3}{c}{Recall (\%)$\uparrow$} \\ \cmidrule(r){1-3} \cmidrule(l){4-6} \cmidrule(l){7-9} \cmidrule(l){10-12}
        Smoothing & WL & MH & Yeast & Filam. & Myco & Yeast & Filam. & Myco & Yeast & Filam. & Myco \\ \midrule
        None & \incorrectmark & \incorrectmark & 67.4 & 37.8 & 72.6 & 63.3 & 41.7 & 75.7 & 27.3 & 22.0 & 38.3 \\
         & \incorrectmark & \correctmark & 68.1 & 37.2 & 73.3 & 57.9 & 38.5 & 78.4 & 26.6 & 20.7 & 34.8 \\
         & \correctmark & \incorrectmark & 72.0 & 41.3 & 75.9 & 63.4 & 41.4 & 77.8 & 31.9 & 25.1 & 46.9 \\
         & \correctmark & \correctmark & 73.1 & 40.0 & 76.0 & 60.3 & 40.1 & 81.4 & 32.4 & 23.9 & 45.0 \\ \midrule
        Standard & \incorrectmark & \incorrectmark & 64.6 & 35.2 & 69.3 & 58.7 & 40.3 & 70.2 & 22.1 & 19.8 & 29.9 \\
         & \incorrectmark & \correctmark & 61.8 & 35.4 & 69.8 & 54.7 & 36.7 & 72.9 & 21.6 & 18.6 & 27.2 \\
         & \correctmark & \incorrectmark & 70.5 & 40.7 & 75.2 & 63.3 & 41.2 & 76.7 & 31.1 & 24.6 & 44.3 \\
         & \correctmark & \correctmark & 72.1 & 39.8 & 75.6 & 61.7 & 40.6 & 80.0 & 31.1 & 23.2 & 42.3 \\ \midrule
        Hierarchical & \incorrectmark & \incorrectmark & 71.8 & 41.3 & 77.5 & 66.6 & 43.0 & 82.7 & 33.0 & 25.4 & 51.8 \\
         & \incorrectmark & \correctmark & 73.3 & 40.3 & 76.5 & 68.9 & 42.5 & \runnerup{83.9} & 33.8 & 24.3 & 48.6 \\
         & \correctmark & \incorrectmark & \runnerup{76.3} & \best{42.6} & \runnerup{78.0} & \runnerup{71.5} & \best{43.9} & 83.1 & \best{39.9} & \best{26.8} & \best{55.5} \\
         & \correctmark & \correctmark & \best{76.8} & \runnerup{42.0} & \best{78.2} & \best{72.0} & \runnerup{43.8} & \best{85.3} & \runnerup{39.2} & \runnerup{26.1} & \runnerup{55.1} \\ \bottomrule
        \end{tabular}
    \end{adjustbox}
    \label{tab:ablation-smoothing}
\end{table}

\subsection{Scaling study}
Using our default configuration (pretrain+fine-tune with BPE tokenizer, hierarchical label smoothing, weighted loss, and multi-head output), we conduct a scaling study (\autoref{fig:model-scaling}). Accuracy on fine-grained ranks (genus and species) is highly sensitive to model capacity. Performance peaks at \textasciitilde 50\,M parameters, consistent with MycoAI findings~\citep[Fig. 8]{mycoai}, then degrades at 140\,M parameters for species-level tasks, suggesting overfitting on fine-grained classification.

\input{figures/model-scaling}
\section{Conclusion}
We addressed fungal DNA barcode classification, a domain with extreme label sparsity and long-tailed distributions.
BarcodeMamba+ demonstrates that SSM-based foundation models using pretrain+fine-tune paradigms substantially outperform fully-supervised approaches. Our systematic evaluation shows BPE tokenization, hierarchical label smoothing, and weighted loss are effective, especially enhancing recall for rare classes.
Our scaling study shows benefits of SSM-based architectures over Transformer-based alternatives while revealing inherent limits on useful model capacity for this task. 

This work enables broader biodiversity research. The enhanced model structure can extend to other genetic markers like COI for insects \citep{elbrecht2019,steinke2024dataset}, and rbcL for plants \citep{cbol2009dna,hollingsworth2011choosing}. We also see opportunities to integrate genomic data with imaging and environmental modalities~\citep[c.f.][]{clibd,Gu2025BioCLIP2}, aligning with growing recognition that comprehensive biodiversity understanding requires diverse data types. By developing scalable AI tools for under-resourced domains like mycology, we can accelerate the pace of species discovery and taxonomic annotation in Earth's most biodiverse yet least understood kingdoms.\looseness=-1

\begin{ack}
BIOSCAN is supported in part by funding from the Government of Canada's New Frontiers in Research Fund (NFRF). Resources used in preparing this research were provided, in part, by the Province of Ontario, the Government of Canada through the Canadian Institute for Advanced Research (CIFAR), and companies sponsoring the Vector Institute \url{http://www.vectorinstitute.ai/\#partners}. AXC and GWT acknowledge support from the Natural Sciences and Engineering Research Council (NSERC), the Canada Research Chairs program, and the Canada CIFAR AI Chairs program.
\end{ack}

\bibliographystyle{iclr2025_conference}
\bibliography{references}

\clearpage
\input{appendix}

\end{document}

%% file: preamble.tex
\usepackage[utf8]{inputenc} %
\usepackage[T1]{fontenc}    %

\usepackage{hyperref}       %
\usepackage{url}            %

\usepackage[dvipsnames,table]{xcolor} %

\usepackage{graphicx}       %
\usepackage{booktabs}       %
\usepackage{multicol}       %
\usepackage{multirow}       %
\usepackage{makecell}       %
\usepackage{colortbl}       %
\usepackage{caption}        %
\usepackage{subcaption}     %
\usepackage{wrapfig}        %
\usepackage{tabularx}       %
\usepackage{enumitem}       %

\usepackage{mathtools}
\usepackage{amsmath,amsthm} %
\usepackage{amsfonts}       %
\usepackage{amssymb}        %
\usepackage{nicefrac}       %
\usepackage{microtype}      %

\usepackage{comment}        %
\usepackage{array}

\usepackage{doi}            %
\usepackage[normalem]{ulem} %
\usepackage{placeins}       %
\usepackage[detect-all]{siunitx} %

\usepackage{cleveref}       %

\usepackage{bbding}         %
\definecolor{cold}{HTML}{008acd}

\usepackage{pifont}         %
\def\cmark{\ding{51}} %
\def\xmark{\ding{55}}
\definecolor{vlightgray}{gray}{0.87}

\def\correctmark{{\color{ForestGreen}\cmark{}}}
\def\incorrectmark{{\color{red}\xmark{}}}

\def\na{{\color{lightgray}N/A}}

\definecolor{tblue}{HTML}{1F77B4}
\definecolor{torange}{HTML}{FF7F0E}
\definecolor{tgreen}{HTML}{2CA02C}
\definecolor{tred}{HTML}{FF0000}

\definecolor{linkcolor}{HTML}{991408}  %
\definecolor{citecolor}{HTML}{2E7E2A}  %
\definecolor{filecolor}{HTML}{131877}  %
\definecolor{menucolor}{HTML}{727500}  %
\definecolor{runcolor} {HTML}{137776}  %
\definecolor{urlcolor} {HTML}{0a2bbf}  %
\hypersetup{colorlinks=true,linkcolor=linkcolor,citecolor=citecolor,filecolor=filecolor,menucolor=menucolor,runcolor=runcolor,urlcolor=urlcolor}

\newlist{compactitem}{itemize}{3}
\setlist[compactitem]{topsep=0pt,partopsep=0pt,itemsep=0pt,parsep=0pt,leftmargin=10pt}
\setlist[compactitem,1]{label=\textbullet}
\setlist[compactitem,2]{label=---}
\setlist[compactitem,3]{label=*}

\newlist{compactdesc}{description}{3}
\setlist[compactdesc]{topsep=0pt,partopsep=0pt,itemsep=0pt,parsep=0pt}

\newlist{compactenum}{enumerate}{3}
\setlist[compactenum]{topsep=0pt,partopsep=0pt,itemsep=0pt,parsep=0pt}
\setlist[compactenum,1]{label=\arabic*}
\setlist[compactenum,2]{label=\alph*}
\setlist[compactenum,3]{label=\roman*}

\newcommand{\appref}[1]{\hyperref[#1]{Appendix~\ref*{#1}}}%

\def\Snospace~{\S{}}%

\newcommand{\best}[1]{\textbf{#1}}
\newcommand{\secbest}[1]{\underline{#1}}
\newcommand{\runnerup}[1]{\secbest{#1}}

\newcommand{\hider}[1]{}

\newcolumntype{Y}{>{\centering\arraybackslash}X}

%% file: figures/model-scaling.tex
\begin{figure}[t]
\centering
\setkeys{Gin}{width=\linewidth}
\includegraphics[trim=10 10 0 10,clip]{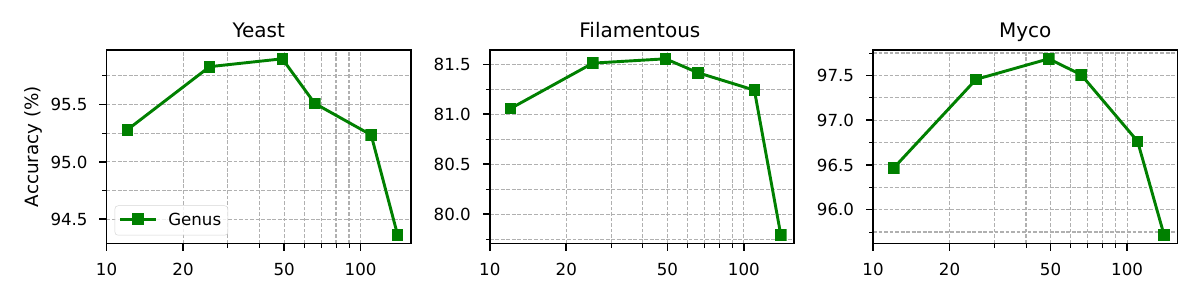}
\includegraphics[trim=10 10 0 10,clip]{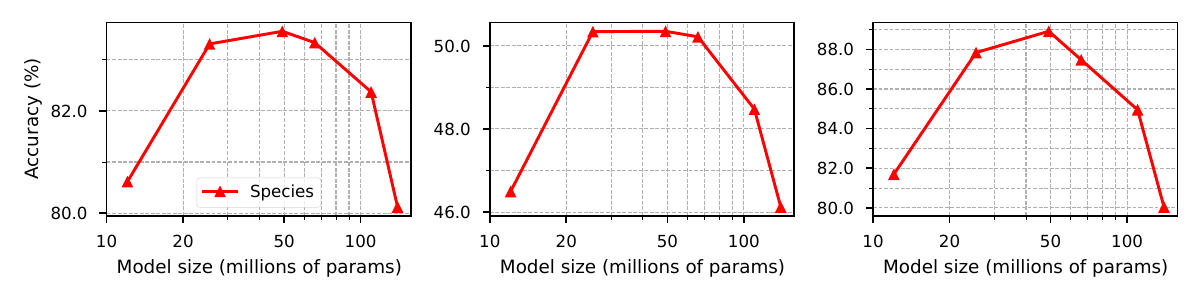}
\caption{Scaling of BarcodeMamba+ to different model sizes. For each test set,  we report the accuracy for classifying at genus (upper panels) and species (lower panels) ranks.}
\label{fig:model-scaling}
\end{figure}

%% file: appendix.tex
\appendix
\section*{Appendices}

\section{Model Architecture Details}
\label{model_architecture_details}

\subsection{Baseline Models}
\label{model_architecture_details_baseline}
\paragraph{Non-learning-based Baseline.}
We use BLASTN~\citep{blast} as a representative non-learning-based method. First, a searchable nucleotide database was constructed from the training set sequences (\texttt{trainset.fasta}) using the \texttt{makeblastdb} command. Sequences from each test set were then aligned against this database using the \texttt{blastn} algorithm. The search was parallelized across 16 CPU threads (\texttt{-num\_threads=16}) for computational efficiency. Results were generated in tabular format (\texttt{-outfmt=6}), providing a list of all significant alignments for each query. In line with exploratory sequence similarity searches, we did not set explicit thresholds for e-value, query coverage, or sequence identity, allowing for the capture of a broad range of potential matches for downstream analysis.

\paragraph{Fully Supervised Baselines.}
We compare against deep learning architectures trained end-to-end without a self-supervised pre-training phase.
\begin{compactitem}
    \item \textbf{CNN Encoder:} This model, introduced by \citet{cnnencoder}, is known for its computational efficiency and accuracy. The architecture consists of three 2D-convolutional layers with kernel sizes of $3 \times 3$, channel dimensions of 64, 32, and 16, respectively, each followed by batch normalization and a ReLU activation function, and interleaved with max-pooling layers of size $3 \times 1$. The final feature maps are flattened and passed through a fully-connected layer.

    \item \textbf{MycoAI-CNN and MycoAI-BERT:} These are the state-of-the-art fully supervised models from \citet{mycoai}. The MycoAI-BERT model is a Transformer-based architecture with 8 encoder layers, 8 attention heads, a hidden dimension of 512, and a feed-forward dimension of 1024. The MycoAI-CNN model is a simple CNN with two convolutional layers (5 and 10 channels, respectively) using a kernel size of 5, followed by max-pooling with pool size 2 and a fully-connected layer of size 256. Both are enhanced by the techniques discussed in \appref{sec:enhancements}. For tokenization, the strongest performing variants were used: BPE for BERT and k-mer-spectral for CNN.
\end{compactitem}

\subsection{Supervised Learning Enhancements for Fine-tuning}
\label{sec:enhancements}
Following \citet{mycoai}, we conduct ablations evaluating three techniques during the fine-tuning stage for both our model and the supervised baselines. 
\begin{compactitem}
    \item \textbf{Hierarchical Label Smoothing (HLS):} Standard label smoothing penalizes confident predictions~\citep{szegedy2016rethinking}. HLS, introduced by \citet{mycoai}, adapts this concept to taxonomy by reducing the penalty for misclassifications that are taxonomically close to the true label (e.g., predicting the correct genus but wrong species). This encourages the model to learn the taxonomic hierarchy.
    \item \textbf{Classification Head:} We compare two output strategies by \citet{mycoai}. The first is a \textbf{multi-head} architecture where separate linear layers predict each of the seven taxonomic ranks simultaneously, allowing the model to learn shared representations. The second is a \textbf{single-head} baseline that predicts only at the species level, with higher-rank probabilities inferred from the species predictions using a pre-defined taxonomic matrix.
    \item \textbf{Weighted Loss:} To counteract the severe class imbalance in the dataset, we adopt the weighted cross-entropy loss from \citet{mycoai}. The loss for each sample is weighted by the inverse square root of its class frequency, encouraging the model to pay more attention to rare taxa.
\end{compactitem}

\subsection{Foundation Model Baseline}
To benchmark our SSM-based approach against the current state-of-the-art in biodiversity foundation models, we include BarcodeBERT~\citep{BarcodeBERT} as our primary comparison point. This transformer-based model was pre-trained on a large-scale invertebrate COI barcode dataset~\citep{canadianinvertebrates} using a masked language modeling objective. It has established strong performance on benchmarks such as the BIOSCAN-5M dataset~\citep{bioscan5m} and is considered an effective architecture for insect biodiversity studies. For our experiments, we use the officially released pre-trained weights and fine-tune the model on our fungal ITS dataset.

\subsection{BarcodeMamba+}
\label{model_architecture_details_barcodemambaplus}
BarcodeMamba+ is a foundation model adapted for the challenges of fungal ITS barcode classification. The model utilizes the BarcodeMamba architecture as its backbone~\citep{barcodemamba}, a powerful SSM previously developed for general DNA sequence analysis. 

\paragraph{Backbone Architecture.}
The BarcodeMamba backbone consists of a stack of \textit{n} identical blocks. Each block processes the input sequence through three main components: a layer normalization step, a multi-layer perceptron, and a Mamba-2 mixing layer. The Mamba-2 layer is the core of the SSM, efficiently capturing long-range dependencies in the DNA sequence by mapping a \textit{d}-dimensional input representation through a \textit{p}-dimensional head. The final hidden states from the backbone serve as rich sequence representations.

\paragraph{Tokenizer.}
To convert raw DNA sequences into input embeddings for the backbone, we evaluated several tokenization strategies. While character-level~\citep{hyenadna} and k-mer-based tokenizers~\citep{BarcodeBERT} have shown success on insect barcode datasets~\citep{barcodemamba}, we integrated a Byte-Pair Encoding (BPE) tokenizer as recommended by \citet{mycoai}. BPE balances between the single-nucleotide resolution of character-level tokens and the pattern-capturing ability of k-mers, while also being vocabulary-efficient and robust to k-mer frameshift issues.

\section{Implementation Details}
\label{sec:implementation-details}
For the BLAST baseline, indexing the training set required 4.6 hours, and classification was performed using a best-hit approach.
For the MycoAI-CNN and MycoAI-BERT models, we followed their official implementation\footnote{\url{https://github.com/MycoAI/MycoAI}}, using the Adam optimizer with 1e-4 weight decay and training for 24 and 16 epochs respectively.
All other models, including the CNN Encoder, BarcodeBERT, and our BarcodeMamba+, were trained using a cross-entropy loss and the AdamW optimizer ($\text{weight\ decay}=0.1$, $\beta_1=0.9$, $\beta_2=0.999$). We used a universal training strategy with an early stopping patience of 3 epochs on the validation loss and a 12-hour time limit.
For our BarcodeMamba+, the fully supervised version was trained for 7 epochs with a learning rate (LR) of 8e-4. In the pre-train/fine-tune paradigm, the model was pre-trained for 15 epochs (LR=8e-4) and subsequently fine-tuned for 12 epochs with a decayed learning rate of 8e-5. 
The BarcodeBERT pretrained model was obtained from HuggingFace\footnote{\url{https://huggingface.co/bioscan-ml/BarcodeBERT}}. The model was fine-tuned with a learning rate of 1e-4 (both as recommended value as reported and the hyperparameter search). Fine-tuning was conducted for 1 epoch. The fully-supervised training process for the CNN Encoder was conducted over 3 epochs. Table \ref{tab:model-settings} summarizes the enhancement settings used for all models.

\begin{table}[tb]
    \centering
    \caption{Optimal settings after hyperparameter search for the comparison study. The reported learning rate is during supervised learning/fine-tuning.}
    \begin{adjustbox}{width=\textwidth}
    \begin{tabular}{@{}llcccl@{}}
    \toprule
     & Label Smoothing & Multi-head & Loss weighting & Learning Rate & Training Strategy \\ \midrule
    BLAST & \na & \na & \na & \na & Index\&Query \\
    MycoAI-CNN (Vu) & Hierarchical & \correctmark & \correctmark & 1e-4 & Fully Supervised \\
    MycoAI-BERT (base) & Hierarchical & \correctmark & \correctmark & 1e-4 & Fully Supervised \\
    CNN Encoder & None & \correctmark & \correctmark & 8e-4 & Fully Supervised \\
    BarcodeBERT & None & \correctmark & \correctmark & 1e-4 & Fine-tuned\\
    BarcodeMamba+ & Hierarchical & \correctmark & \correctmark & 8e-5 & Pretrained, Fine-tuned\\
    BarcodeMamba+ (large) & Hierarchical & \correctmark & \correctmark & 8e-5 & Pretrained, Fine-tuned\\ \bottomrule
    \end{tabular}
    \end{adjustbox}
    \label{tab:model-settings}
    \end{table}

\section{Statistical Tests}
\label{s:statistical-tests}

To establish whether ablated model components had a significant effect on the model performance, we conducted a series of paired $t$-tests for each component.
We assumed each component would have an independent effect on the performance, and compared the accuracy, precision, or recall with one component present versus ablated, across all other component configurations.
The results are shown in \autoref{tab:ablation-significance}.

\begin{table}[tb]
    \centering
    \caption{Statistical significance ($p$-values) of different model components on species-level metrics across three test sets. Both label smoothing types are compared against a baseline of no label smoothing. Significant results ($p < 0.05$) on a paired $t$-test are highlighted in bold.}
    \begin{tabular}{ll rr rr rr}
        \toprule
        Component & Test Set & Accuracy & Precision & Recall \\
        \midrule
        
        Weighted loss & Yeast & \textbf{0.002} & \textbf{0.013} & \textbf{< 0.001} \\
        & Filamentous & \textbf{0.004} & 0.052 & \textbf{0.003} \\
        & MycoAI & \textbf{0.014} & \textbf{0.030} & \textbf{0.003} \\
        \addlinespace
       
        Multi-head & Yeast & 0.550 & 0.173 & 0.762 \\
        & Filamentous & \textbf{0.018} & 0.053 & \textbf{< 0.001} \\
        & MycoAI & 0.626 & \textbf{< 0.001} & \textbf{0.004} \\
        \addlinespace
       
        Hierarchical label smoothing & Yeast & \textbf{< 0.001} & \textbf{0.021} & \textbf{< 0.001} \\
        & Filamentous & \textbf{0.017} & \textbf{0.020} & \textbf{0.010} \\
        & MycoAI & \textbf{0.019} & \textbf{0.004} & \textbf{0.003} \\
        \addlinespace
       
        Standard label smoothing & Yeast & 0.093 & 0.315 & 0.081\\
        & Filamentous & 0.091 & 0.266 & 0.056 \\
        & MycoAI & 0.095 & 0.070 & \textbf{0.042} \\
        \bottomrule
        \end{tabular}
    \label{tab:ablation-significance}
\end{table}

\section{Dataset Analysis}
\label{s:dataset-analysis}

In \autoref{tab:dataset_overlap}, both Test Set 1 (Yeast) and Test Set 3 (MycoAI Benchmark) have high identical barcode overlap with the training data, at 86.73\% and 100.00\% respectively. This shows that they primarily measure model performance on in-distribution sequences. Conversely, Test Set 2 (Filamentous Fungi) shows minimal overlap at only 6.48\%, establishing it as the most rigorous and reliable benchmark in this study for evaluating generalization capabilities on unseen species and barcodes.

\begin{table}[h!]
\centering
\caption{Analysis of dataset overlap between the training set and the three test sets. Overlap percentages are calculated relative to the unique species or barcodes in each test set, respectively.}
\label{tab:dataset_overlap}
\resizebox{\columnwidth}{!}{
\begin{tabular}{l rrrc rrr}
\toprule
& \multicolumn{3}{c}{\textbf{Species Overlap}} & & \multicolumn{3}{c}{\textbf{Identical Barcode Overlap}} \\
\cmidrule(lr){2-4} \cmidrule(lr){6-8}
\textbf{Test Set} & \textbf{Total (Test)} & \textbf{Overlap (n)} & \textbf{Overlap (\%)} & & \textbf{Total (Test)} & \textbf{Overlap (n)} & \textbf{Overlap (\%)} \\
\midrule
\textbf{Test Set 1: Yeast} & 1,157 & 616 & 53.24\% & & 4,235 & 3,673 & 86.73\% \\
\textbf{Test Set 2: F. Fungi} & 5,537 & 2,650 & 47.86\% & & 10,721 & 695 & 6.48\% \\
\textbf{Test Set 3: MycoAI} & 14,742 & 14,742 & 100.00\% & & 363,420 & 363,420 & 100.00\% \\
\bottomrule
\end{tabular}
}
\end{table}